\documentclass[11pt]{article}

\usepackage{enumitem}
\usepackage{amsmath}
\usepackage{xcolor}
\usepackage{colortbl}
\usepackage{graphicx}

\usepackage[preprint]{acl}

\usepackage{times}
\usepackage{latexsym}

\usepackage[T1]{fontenc}

\usepackage[utf8]{inputenc}

\usepackage{microtype}

\usepackage{inconsolata}

\usepackage{graphicx}

\usepackage{xcolor}
\usepackage{lmodern}
\usepackage{graphicx}
\usepackage[svgnames]{xcolor}

\newcommand{\PrismAgent}{%
  {\textcolor{red!80!blue}{P}
   \textcolor{orange!70!yellow}{r}
   \textcolor{orange!90!white}{i}
   \textcolor{green!65!orange}{s}
   \textcolor{cyan!60!white}{m}
   \textcolor{SkyBlue!65!black}{A}
   \textcolor{blue!60!white}{g}
   \textcolor{violet!65!white}{e}
   \textcolor{magenta!60!orange}{n}
   \textcolor{violet!80!white}{t}}
}

\title{\raisebox{-0.7em}{\includegraphics[height=2.0em]{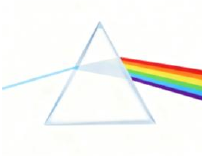}}\hspace{0.2em}\textbf{\PrismAgent}: Illuminating Harm in Memes via a Zero-Shot Interpretable Multi-Agent Framework}

\author{
  Zihan Ding\textsuperscript{1}
  \and
  Ziyuan Yang\textsuperscript{2}
  \and
  Yi Zhang\textsuperscript{1}\thanks{*Corresponding author: \href{mailto:yzhang@scu.edu.cn}{yzhang@scu.edu.cn}}
  \\[6pt] 
  \textsuperscript{1}Sichuan University, Chengdu, Sichuan, China \\
  \texttt{2022141230168@stu.scu.edu.cn}, \texttt{yzhang@scu.edu.cn}
  \\[4pt]
  \textsuperscript{2}Nanyang Technological University, Singapore \\
  \texttt{cziyuanyang@gmail.com}
}

\begin{document}
\maketitle

\begin{abstract}
The rapid spread of memes makes harmful content detection increasingly crucial, as effective identification can curb the circulation of misinformation.
However, existing methods rely heavily on high-volume annotated data, which leads to substantial training costs and limited generalization.
To address these challenges, we propose \textbf{PrismAgent}, a zero-shot, multi-agent, interpretable framework.
PrismAgent conceptualizes this task as a criminal case investigation, employing four specialized agents responsible for the analysis, investigation, prosecution, and judgment stages within a structured collaborative workflow.
In the first stage, the analyst agent paraphrases each meme under benevolent and malicious assumptions to probe its underlying intent.
The investigator agent then retrieves supporting evidence from an unannotated dataset and constructs contextual interpretations for the meme and its variants.
Next, the prosecutor agent performs three independent preliminary judgments by pairing the original meme with each of the three interpretations.
Finally, the judge agent deliberates across all evidence to render a final verdict.
Moreover, PrismAgent's explicit multi-stage reasoning chain makes the model inherently interpretable, since every intermediate step is explicitly explained rather than only producing a final detection result.
Extensive experiments on three public datasets show that PrismAgent significantly outperforms existing zero-shot detection methods.
\end{abstract}

\section{Introduction}

With the rapid development of social media, memes have become a pervasive form of multimodal communication. While memes were initially created for humor expression, they are increasingly being misused to spread misinformation and harmful biases. This emerging trend poses serious risks to social media, highlighting the necessity of detecting harmful memes~\citep{Sharma_2022}.

Traditional harmful meme detection methods primarily rely on large-scale and carefully annotated dataset~\citep{Yang_Yan_Chen_Wang_Lu_Zhang_2024,Lin_2024}.
However, these methods tend to overfit to the training distribution, which compromises their robustness when applied to unseen or evolving memes. Given the rapidly changing nature of meme content, such approaches face substantial challenges when deployed in real-world scenarios~\citep{Cao_2024,Huang_2024}.

Another challenge lies in the inherently implicit nature of memes. Their meaning is often conveyed through shared cultural knowledge, visual symbolism, or multimodal irony rather than explicit textual content. Such indirect expressions make it challenging for models to capture the underlying intent, as understanding memes typically requires commonsense reasoning and cultural context that extend beyond literal interpretation.

To address these challenges, we propose \textbf{PrismAgent}, a zero-shot, multi-agent, interpretable framework for harmful meme detection.
Much like a prism that refracts a single beam of light into multiple revealing facets, PrismAgent decomposes each meme into diverse motivational and contextual perspectives, which makes its underlying meaning clearer and its judgment more interpretable. To achieve this, PrismAgent simulates the workflow of a real criminal case investigation, where different stages of the process are handled by distinct specialized roles. The investigation begins with analysts who infer potential motives and intentions behind a meme. Investigators then collect relevant contextual evidence to reason about the case and establish preliminary interpretations. Finally, a judge evaluates both the collected evidence and the investigators’ reasoning to deliver a fair and well-grounded verdict.

\begin{figure}[!t]
  \centering
  \includegraphics[width=\linewidth]{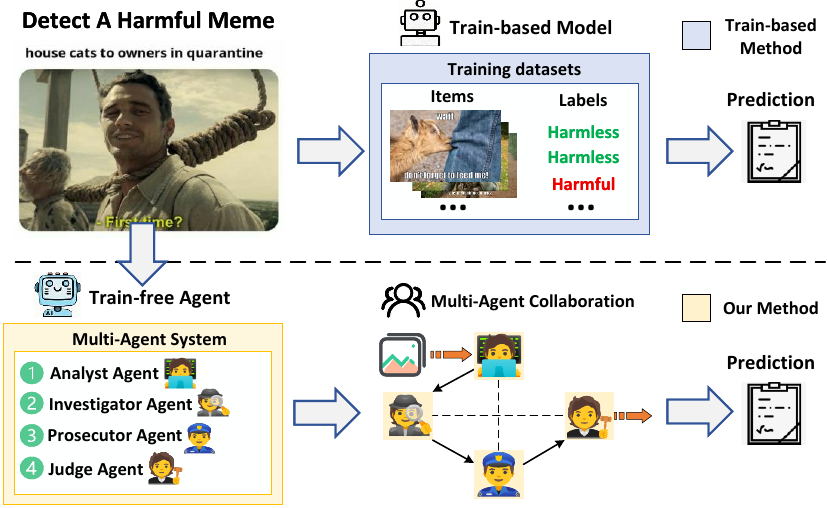}
  \vspace{-15pt}
  \caption{Pipeline comparison between training-based methods and our proposed approach.}
  \vspace{-15pt}
  \label{fig:introduce}
\end{figure}

PrismAgent instantiates each stage of the investigation process as a dedicated agent, including the analyst, investigator, prosecutor, and judge. The analyst agent paraphrases each meme under two opposing disseminator intentions, benevolent and malicious, to reveal the range of plausible potential intent. Next, the investigator agent searches the unannotated dataset for evidence relevant to the original meme and its two rewritten variants, which are then used to construct contextual interpretations. The prosecutor agent then reasons over the original meme and each contextualized variant, producing three independent judgments.
If these judgments are consistent, the prosecutor finalizes the decision. Otherwise, the case is forwarded to the judge agent, who integrates all available information and deliberates to render a coherent and interpretable final verdict.

Unlike previous methods that rely on a single-step prediction, we decompose harmful meme detection into a series of structured subtasks. This design induces an explicit reasoning chain and substantially improves the interpretability of the detection process. Moreover, by adopting a reasoning-based pipeline rather than training a task-specific classifier, PrismAgent eliminates the dependence on annotated data and naturally enables effective zero-shot detection. A comparison between previous methods and ours is illustrated in Fig.~\ref{fig:introduce}. Our main contributions are summarized as follows:


\begin{itemize}
\item We propose PrismAgent, a multi-agent framework that decomposes harmful meme detection into structured subtasks, which enables modular role-specific reasoning.
\item We design a multi-stage reasoning chain that integrates the outputs of several specialized agents to enhance interpretability via explicit, stage-wise explanations.
\item Extensive experiments on public datasets demonstrate the effectiveness of PrismAgent and show that it performs competitively against state-of-the-art approaches. 
\end{itemize}

\begin{figure*}[!t]
  \centering
  \includegraphics[width=\linewidth]{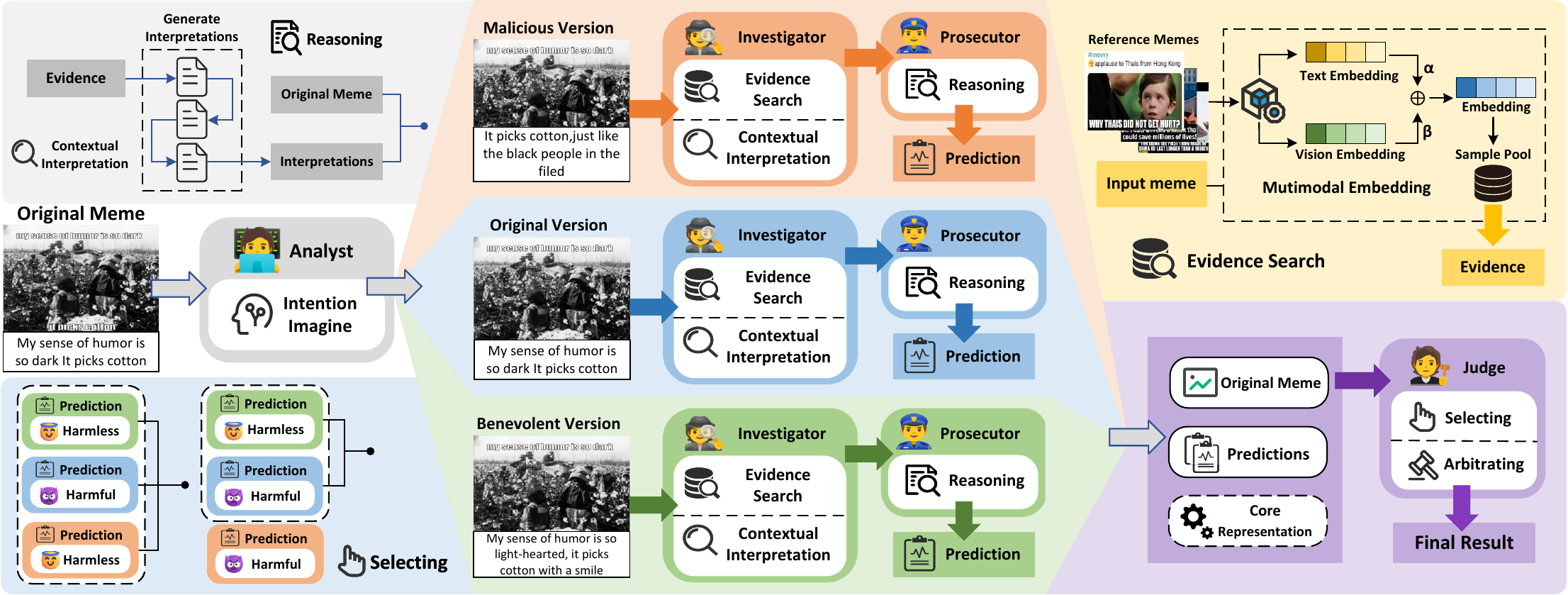}
  \caption{Overview of the proposed PrismAgent.}
  \vspace{-15pt}
  \label{fig:overview}
\end{figure*}

\section{Related Works}

\textbf{Harmful Meme Detection}
The multimodal nature of memes necessitates multimodal strategies \citep{Borakati_2021,Beyer_Alexy_2025}, which consistently outperform unimodal methods in harmful meme detection \citep{He_2016,Devlin_2019}. Previous studies primarily follow two methodological directions. The first relies on classical two-stream architectures that perform harmful meme classification by fusing textual and visual modalities \citep{Kiela_2022,Suryawanshi_2020,Pramanick_2021,kumar_2022}. The second direction fine-tunes large pre-trained multimodal models to adapt them to domain-specific detection tasks \citep{Velioglu_Rose_2020,Hee_2022,Hee_2025,Hee_Lee_2025}. In addition, relevant studies have proposed various improved strategies \citep{Cao_2022,Ji_2023,Cao_2023,Garg_2025,Kumari_2025}, as well as detection methods for few-shot annotation \citep{Cao_2024, Huang_2024} and zero-shot annotation \citep{liu-etal-2025-mind} scenarios to reduce reliance on labeled data.

\noindent \textbf{VLM-based multi-agent framework}
When Vision Language models~(VLMs) are deployed as agents across various domains, they demonstrate strong planning and reasoning capabilities in a wide range of scenarios \citep{Yao_2022,Sun_2023,Bang_2024_CVPR,Zhao_2024,Yang_2024_CVPR,Chen_2024_CVPR,gao_2025}. These advancements show that LLM-based methods can effectively handle complex tasks even under weakly supervised conditions. Building on the success of single-agent systems, multi-agent frameworks \citep{Qian_2024,Tao_2024,Hong_2024_CVPR,Ma_2024_CVPR,Zahedifar_2025,yang2025multiagentvisuallanguagereasoningcomprehensive} further enable interactive collaboration and collective problem-solving. However, research on multi-agent frameworks for harmful meme detection remains limited. Although \citet{liu-etal-2025-mind} proposed the MIND framework, it fails to fully address the challenge of interpreting meme metaphors under zero-shot settings.

\section{Method}
\subsection{Overview}

Detecting harmful memes presents a significant challenge because their meaning often relies on subtle cultural references, visual symbols, or humorous wordplay, rather than explicit textual or visual cues. Additionally, most existing methods depend on large-scale annotated training datasets, which often cause detectors to overfit to the training memes. As memes continually evolve, these methods struggle with poor generalization.


To alleviate these challenges, we propose PrismAgent, a zero-shot, multi-agent, interpretable framework for harmful meme detection. An overview of the framework is depicted in Figure~\ref{fig:overview}. PrismAgent introduces an explicit reasoning chain and assigns distinct roles to four agents, including analyst, investigator, prosecutor, and judge. These agents collaborate to reveal the intentions behind memes. Specifically, the analyst agent imagines two opposing intentions for the meme's disseminator and paraphrases the meme accordingly. The investigator agent then searches an unannotated dataset for relevant supporting evidence for the original meme and its two rewrites, which construct contextual interpretations for each version. The prosecutor agent produces three independent judgments by reasoning over the original meme together with each of the contextual interpretations.
If these judgments conflict, the judge agent integrates all evidence and delivers a final, robust, and interpretable decision.

PrismAgent requires no training and operates as a zero-shot detection framework. By leveraging a structured multi-agent reasoning process, it renders decisions that are highly interpretable and transparent.

\subsection{Analyst Agent}
Revealing the hidden intent of memes is a crucial yet challenge step in meme understanding. To make such implicit intentions more accessible, we introduce the analyst agent to explicitly amplify implicit intentions by assuming two opposing disseminator intentions, benevolent and malicious, to reveal the real intent.

Specifically, for each meme version, we first input it into the agent, and the agent is guided by the carefully designed prompts to imagine possible dissemination motives and paraphrase two corresponding rewrites: one reflecting benevolent and the other reflecting malicious intensions. 
In this process, the original linguistic style and contextual tone are preserved as much as possible, ensuring that the rewritten memes remain semantically coherent while exposing their potential intentions. Consequently, the procedure produces two rewritten versions that make the original meme’s latent intentions more explicit from two opposing motives. This process can be expressed as:
\begin{equation}
M_b = \textit{Agent}_{Ana}(M_{ori},P_b),
\end{equation}
\begin{equation}
M_m = \textit{Agent}_{Ana}(M_{ori},P_m),
\end{equation}
where \(M_{ori} = \{\mathcal{V}, \mathcal{T}\}\) denotes the original meme. \(M_b = \{\mathcal{V}, \mathcal{T}_b\}\) and \(M_m = \{\mathcal{V}, \mathcal{T}_m\}\) represent the benevolent and malicious memes obtained by our analyst agent, respectively. $\mathcal{V}$ and $\mathcal{T}$ denote the visual content and the accompanying text, respectively.
\(\mathcal{T}_b\) and \(\mathcal{T}_m\) represent the benevolent and malicious rewrites, respectively. Besides, $P_b$ and $P_m$ denote the prompts used to rewrite the meme from benevolent and malicious intentions, respectively.

Finally, a single original meme \(M_{ori}\) is expanded into a set \(\{M_{ori}, M_b, M_m\}\), which contains the original sample and its two rewrites. This process makes the latent intent more salient by forcing the meme to be expressed under contrasting motivational assumptions, thereby amplifying the subtle cues embedded in the original content. The detailed prompts are provided in Appendix \ref{sec:appendix_IIE}.

\subsection{Investigator Agent}
After the analyst agent paraphrases each meme, the investigator agent then searches the unannotated dataset for supporting evidence to the original meme and its two rewrites. This step provides the subsequent investigation stage with additional context and precedents, enabling it to gather supporting evidence that further explores the potential intentions behind the original meme and its rewrites. Moreover, this process offers a potential advantage: since memes are constantly evolving, the investigator agent can continuously expand the unannotated dataset in the deployment stage to incorporate emerging contexts. This allows PrismAgent to adapt to emerging meme trends in real time, enhancing its robustness and generalization in the zero-shot scenario.
The evidence search process can be formulated as:
\begin{equation}
\mathcal{Q}_{\text{evi}}=f(D_{ref},M_{ori})
\end{equation}
where $\mathcal{Q}_{\text{evi}}=\{M_\text{evi}^1,...,M_\text{evi}^k$\} denotes the set of top-$k$ relevant memes retrieved from the unannotated reference dataset $D_{ref}$. $M_\text{evi}^k$ denotes most $k$-th most similar meme retrieved from $D_{ref}$. $f(\cdot)$ represents the matching function that measures the similarity between the original meme and each sample in $D_{ref}$ and returns the top-$k$ closest memes. $k$ is empirically set to 3 in this paper.

After retrieving the relevant cases, the investigator agent interprets them in descending order of similarity. At each step, the current meme is interpreted together with the interpretations accumulated from previous steps, enabling the agent to integrate newly evidence with prior observations. This sequential accumulation of information helps the investigator agent to capture subtle contextual cues and latent intentions that may not be apparent from a single meme. This process can be formulated as follows:
\begin{equation}
\mathcal{O}_{i} = \textit{Agent}_{Inv} (M^i_{\text{evi}}, \mathcal{O}_{i-1}, P_{int} ),
\end{equation}
where \(\mathcal{O}_i\) and \(\mathcal{O}_{i-1}\) represent the interpretations generated in the $i$-th and ($i$-1)-th steps, respectively; $M^i_{\text{evi}}$ denotes the $i$-th most similar evidence retrieved from $D_{ref}$. $P_{int}$ denotes the interpretation prompt, which is detailed in Appendix \ref{sec:appendix_CC&IE}.

\subsection{Prosecutor Agent}
\label{sec:proagent}
After the investigator agent has gathered relevant evidence and constructed contextual interpretations for the original meme and its rewrites, the prosecutor agent takes over to perform harm assessment. The prosecutor agent evaluates the original meme with different interpretations individually and issues an independent judgment regarding its potential harmfulness. In this way, the prosecutor agent reasons the harmfulness based on different assumed intentions and detect whether harmful meaning persists across  perspectives, thereby providing a more reliable basis for final judgment. The process can be formulated as:
\begin{equation}
\small
\mathcal{R}_u = \textit{Agent}_{Pro}(\mathcal{O}_u, M_{ori},P_\text{pro}),\quad
u\in\{ori,b,m\},
\end{equation}
where $\mathcal{R}$ denotes the prosecution result, denotes the prosecution result, which includes both the harmful judgment and the supporting rationale. $P_\text{pro}$ denotes the prosecution prompt, which is detailed in Appendix~\ref{sec:appendix_prosecutor}.

The interpretations captured by the investigator agent reflect how the meme’s harmfulness manifests under distinct dissemination intents. The prosecutor agent reasons over these interpretations to assess whether the harmful interpretation shifts when the meme is rewritten with benevolent or malicious motives. Accordingly, if the harmful judgments from $\mathcal{R}_{\text{ori}}, \mathcal{R}_{\text{b}}, \mathcal{R}_{\text{m}}$ converge to the same conclusion, then no further arbitration is required.

\subsection{Judge Agent}
During arbitration by the judge agent, we guide the model to concentrate on the core points of contention across different viewpoints. To enhance the objectivity and reliability of the analysis, we introduce auxiliary evidence into the arbitration process.
Specifically, we reuse $\mathcal{Q}_\text{evi}$ as the input of the judge agent. Unlike the stepwise interpretation of similar samples mentioned in Sec.~\ref{sec:proagent}, here the original meme and its evidence are fed into the judge agent simultaneously. This enables the judge agent to conduct a joint analysis that takes both the original meme and its evidence into account to reason from a broader contextual basis, which can be formally expressed as:
\begin{equation}
\mathcal{R}_\text{jud}\ = \textit{Agent}_{Jud}(M_{ori}, \mathcal{Q}_{\text{evi}},P_\text{core}),
\end{equation}
where \(\mathcal{R}_\text{jud}\) represents the core representation, $P_\text{core}$ denotes the related prompt.

\(\mathcal{R}_\text{jud}\) summarizes the core characteristics shared by similar memes, such as common themes (e.g., pandemics, racial issues) and expressive techniques (e.g., exaggeration, irony). During arbitration, it serves as an anchor that guides the agent toward relevant reasoning paths, preventing irrelevant reasoning and improving the objectivity and robustness of the decision. The judge agent integrates all information and complete arbitration process as:
\begin{equation}
\small
    \mathcal{J} = \textit{Agent}_{Jud}(\mathcal{R}_\text{ori},\mathcal{R}_\text{amb},{M}_{\text{ori}}, \mathcal{R}_\text{jud}\ ,P_{jud}),
\end{equation}
where $\mathcal{R}_{\text{amb}}$ denotes the ambiguity prosecution result, which is inconsistent with the original one. $\mathcal{R}_{\text{amb}} = \{ \mathcal{R}_{\text{x}} \mid \mathcal{R}_{\text{x}} \neq \mathcal{R}_{\text{ori}},\ x \in \{b, m\} \} $. $P_{jud}$ denotes the final judge prompt. $\mathcal{J} $ denotes the final judgment result and the related supporting rationale. The detailed prompts are provided in Appendix \ref{sec:appendix_judge}.

In this way, we construct a reasoning chain that mirrors the practical workflow of a criminal case investigation. Hence, PrismAgent is an interpretable framework that provides a stage-wise reasoning process supported by explicit reasoning interpretations, rather than merely outputting a final detection result.

\begin{table*}[ht]
\footnotesize
\setlength{\tabcolsep}{6pt}  
\renewcommand{\arraystretch}{1.2}
\resizebox{\linewidth}{!}{
\begin{tabular}{l||cc||cc||cc}
\hline
Dataset & \multicolumn{2}{c||}{HarM} & \multicolumn{2}{c||}{FHM} & \multicolumn{2}{c}{MAMI} 
\\
\hline
Model & Accuracy & Macro-$F_1$ & Accuracy & Macro-$F_1$ & Accuracy & Macro-$F_1$ 
\\ \hline\hline
\multicolumn{7}{c}{\textbf{Supervised / Training-Based Methods}}
\\ \hline
Late Fusion \citep{Pramanick_2021a} & 73.24 & 70.25 & 59.14 & 44.81 & 63.20 & 59.76
\\
MOMENTA \citep{Pramanick_2021} & \textbf{83.32} & \textbf{82.80} & 61.34 & 57.45 & 72.10 & 66.93
\\ \hline\hline
\multicolumn{7}{c}{\textbf{Zero-Shot Methods}}
\\ \hline
\multicolumn{7}{l}{\textit{Closed-Source VLMs}}
\\ \hline
GPT-4o \citep{Achiam_2023} & 67.51 & 60.29 & \textbf{68.80} & \textbf{68.25} & \textbf{81.00} & \textbf{81.00}
\\
Gemini-2.0-Flash \citep{Team_2024} & 64.67 & 58.84 & 60.40 & 54.04 & 80.00 & 79.92
\\ \hline
\multicolumn{7}{l}{\textit{Open-Source VLMs}}
\\ \hline
LLaVA-1.5-7B \citep{Liu_2024} & 59.23 & 49.44 & 53.80 & 45.51 & 52.90 & 41.53 
\\
InstructBLIP-7B \citep{Dai_2023} & 51.53 & 50.99 & 52.00 & 48.85 & 53.10 & 46.93 
\\
MiniGPT-V-2B \citep{Chen_2023} & 60.12 & 52.39 & 51.30 & 47.88 & 57.40 & 52.22 
\\
OpenFlamingo-9B \citep{Awadalla_2023} & 63.42 & 54.36 & 50.50 & 49.52 & 54.70 & 49.88
\\
LLaVA-1.5-13B \citep{Liu_2024} & 62.28 & 50.45 & 55.20 & 53.01 & 60.10 & 55.52 
\\
InstructBLIP-13B \citep{Dai_2023} & 64.92 & 49.61 & 55.40 & 51.89 & 60.00 & 57.97
\\
LLaVA-1.6-34B \citep{Liu_2024} & 67.51 & 61.59 & 64.00 & 63.51 & 71.30 & 71.28
\\ \hline
\multicolumn{7}{l}{\textit{Agent-Based Methods (Open-Source)}}
\\ \hline
MIND (LLaVA-1.5-13B) \citep{liu-etal-2025-mind} & 68.93 & 65.19 & 60.80 & 60.71 & 68.90 & 68.84
\\
\rowcolor{gray!15} PrismAgent (LLaVA-1.5-13B) & 70.62 & 68.44 & 64.00 & 63.96 & 70.70  & 70.69
\\
\rowcolor{gray!15} PrismAgent (LLaVA-1.6-34B) & \underline{71.19} & \underline{69.78} & \underline{66.80} & \underline{66.72} & \underline{73.20}  & \underline{73.18}
\\
\hline
\end{tabular}
}
\vspace{-5pt}
\caption{Harmful meme detection results on three datasets. \textbf{Bold} indicates the absolute best performance. \underline{Underline} denotes the best performance strictly within the open-source zero-shot methods.}
\vspace{-5pt}
\label{tab:model_performance}
\end{table*}

\begin{table*}[ht]
\scriptsize
\setlength{\tabcolsep}{4pt}
\renewcommand{\arraystretch}{1.2}
\resizebox{\linewidth}{!}{
\begin{tabular}{l||cc||cc||cc}
\hline
Dataset & \multicolumn{2}{c|}{HarM} & \multicolumn{2}{c|}{FHM} & \multicolumn{2}{c}{MAMI} 
\\
\hline
Model & Accuracy & Macro-$F_1$ & Accuracy & Macro-$F_1$ & Accuracy & Macro-$F_1$ 
\\
\hline
LLaVA-1.5-7B & 59.23 & 49.44 & 53.80 & 45.51 & 52.90 & 41.53 
\\
MIND (LLaVA-1.5-7B) & 62.71 \textcolor{red}{(+3.48)} & 57.22 \textcolor{red}{(+7.78)} & 54.00 \textcolor{red}{(+0.20)} & 48.28 \textcolor{red}{(+2.77)} & 53.90 \textcolor{red}{(+1.00)} & 45.45 \textcolor{red}{(+3.92)}
\\
\rowcolor{gray!15} PrismAgent (LLaVA-1.5-7B) & 63.56 \textcolor{red}{(+4.33)} & 58.59 \textcolor{red}{(+9.15)} & 53.80 \textcolor{red}{(+0.00)} & 49.29 \textcolor{red}{(+3.78)} & 55.10 \textcolor{red}{(+2.20)} & 47.48 \textcolor{red}{(+5.95)}
\\ \hline
LLaVA-1.5-13B & 62.28 & 50.45 & 55.20 & 53.01 & 60.10 & 55.52 
\\
MIND (LLaVA-1.5-13B) & 68.93 \textcolor{red}{(+6.65)} & 65.19 \textcolor{red}{(+14.47)} & 60.80 \textcolor{red}{(+5.60)} & 60.71 \textcolor{red}{(+7.70)} & 68.90 \textcolor{red}{(+8.80)} & 68.84 \textcolor{red}{(+13.28)}
\\ 
\rowcolor{gray!15} PrismAgent (LLaVA-1.5-13B) & 70.62 \textcolor{red}{(+8.34)} & 68.44 \textcolor{red}{(+17.99)} & 64.00 \textcolor{red}{(+8.80)} & 63.96 \textcolor{red}{(+10.95)} & 70.70 \textcolor{red}{(+10.60)} & 70.69 \textcolor{red}{(+15.17)}
\\ \hline
LLaVA-1.6-34B & 67.51 & 61.59 & 64.00 & 63.51 & 71.30 & 71.28 
\\
MIND (LLaVA-1.6-34B) & 69.49 \textcolor{red}{(+1.98)} & 66.12 \textcolor{red}{(+4.53)} & 66.40 \textcolor{red}{(+2.40)} & 68.38 \textcolor{red}{(+4.87)} & 73.60 \textcolor{red}{(+2.30)} & 75.38 \textcolor{red}{(+4.10)}
\\
\rowcolor{gray!15} PrismAgent (LLaVA-1.6-34B) & 71.19 \textcolor{red}{(+3.68)} & 69.78 \textcolor{red}{(+8.19)} & 66.80 \textcolor{red}{(+2.80)} & 66.72 \textcolor{red}{(+3.21)} & 73.20 \textcolor{red}{(+1.90)} & 73.18 \textcolor{red}{(+1.90)}
\\ \hline
Gemini-2.0-FLash & 64.67 & 58.44 & 60.40 & 54.04 & 80.00 & 79.92
\\
MIND (Gemini-2.0-FLash) & 64.84 \textcolor{red}{(+0.17)} & 60.13 \textcolor{red}{(+1.69)} & 67.20 \textcolor{red}{(+6.80)} & 66.05 \textcolor{red}{(+12.01)} & 80.10 \textcolor{red}{(+0.10)} & 80.00 \textcolor{red}{(+0.08)}
\\
\rowcolor{gray!15} PrismAgent (Gemini-2.0-FLash) & 67.51 \textcolor{red}{(+2.84)} & 63.05 \textcolor{red}{(+4.61)} & 68.20 \textcolor{red}{(+7.80)} & 67.51 \textcolor{red}{(+13.47)} & 80.50 \textcolor{red}{(+0.50)} & 80.46 \textcolor{red}{(+0.54)}
\\
\hline
\end{tabular}
}
\vspace{-5pt}
\caption{Performance improvements of PrismAgent across different model scales and datasets for zero-shot harmful meme detection. Numbers in \textcolor{red}{red} indicate absolute improvements over original models.}
\vspace{-10pt}
\label{tab:accuracy_improvements}
\end{table*}
\section{Experiment}
\subsection{Experiment Setup}
\textbf{Datasets.} We use three publicly available meme datasets for evaluation: (1) HarM \citep{Pramanick_2021}, (2) FHM \citep{Kiela_2020}, and (3) MAMI \citep{Fersini_2022}. 

\noindent \textbf{Baselines.} We compare PrismAgent with various advanced methods: 1) GPT-4o \citep{Achiam_2023}; 2) Gemini-2.0-Flash \citep{Team_2024} 3) Late Fusion \citep{Pramanick_2021a}; 4) MOMENTA \citep{Pramanick_2021} 5) LLaVA-1.5-7B \citep{Liu_2024}; 6) InstructBLIP-7B \citep{Dai_2023}; 7) MiniGPT-v2-7B \citep{Chen_2023}; 8) OpenFlamingo-9B \citep{Awadalla_2023}; 9) LLaVA-1.5-13B \citep{Liu_2024}; 10) InstructBLIP-13B \citep{Dai_2023}; 11) LLaVA-1.5-34B \citep{Liu_2024}; 12) MIND \citep{liu-etal-2025-mind}. 

\noindent \textbf{Metrics.} We use the accuracy and macro-averaged F1 scores as the evaluation metrics. 
The detail experimental setting and implementation details can be found in Appendix \ref{sec:appendix_Implementation_Details}.

\subsection{Harmful Meme Detection Experiments}
We evaluate PrismAgent against a range of existing methods on three public datasets, and the results are summarized in Table~\ref{tab:model_performance}. These methods can be roughly divided into three groups, including the training-based methods, zero-shot methods and agent-based methods. It can be seen that PrismAgent consistently outperforms other baselines, demonstrating its strong detection performance. Even with a 13B backbone, PrismAgent achieves performance comparable to models with over 34B parameters, highlighting its efficiency and reasoning capability. 
Compared with the latest detection framework MIND, it also achieves improvements of 2.23\% in average accuracy and 2.78\% in macro F1-score, fully demonstrating its performance advantages. Moreover, we investigate the robustness and computational overhead experiments, and the results are shown in Appendix \ref{appendix:robustness} and \ref{appendix:computational overhead}.

\subsection{Generalization Experiments}

To verify the generalization of PrismAgent, we integrate it with various VLMs, including both open- and closed-source models. Besides, we compare our performance with MIND. As shown in Table~\ref{tab:accuracy_improvements}, PrismAgent consistently enhances performance across the majority of models and datasets, demonstrating its adaptability to diverse backbones. Among them, the performance improvement on LLaVA-1.5-13B is particularly remarkable; even for the more powerful LLaVA-1.6-34B model, the average accuracy and macro F1-score across three datasets are increased by 2.79\% and 4.43\%, respectively.

\subsection{Ablation Study}
\begin{table*}[ht]
\small
\centering
\renewcommand{\arraystretch}{1.2}
\begin{tabular}{l||cc|cc|cc}
\hline
Dataset & \multicolumn{2}{c|}{HarM} & \multicolumn{2}{c|}{FHM} & \multicolumn{2}{c}{MAMI} \\
Model & Accuracy & Macro-$F_1$ & Accuracy & Macro-$F_1$ & Accuracy & Macro-$F_1$ \\
\hline
Baseline & 62.28 & 50.45 & 55.20 & 53.01 & 60.10 & 55.52 \\
PrismAgent (LLaVA-1.5-13B) & 70.62 & 68.44 & 64.00 & 63.96 & 70.70 & 70.69 \\
\quad w/o Analyst Agent & 67.80 & 59.00 & 60.00 & 59.49 & 67.30 & 66.09 \\
\quad w/o Investigator Agent & 66.67 & 60.49 & 58.20 & 57.65 & 63.90 & 62.53 \\
\quad w/o Prosecutor Agent & 60.73 & 53.58 & 52.20 & 52.13 & 60.10 & 59.96 \\
\quad w/o Judge Agent & 67.80 & 65.13 & 61.60 & 61.35 &  68.20 & 68.03 \\
\hline
\end{tabular}
\vspace{-5pt}
\caption{Ablation studies on our proposed framework.}
\vspace{-10pt}
\label{tab:ablation}
\end{table*}
To evaluate the effectiveness of different agents in PrismAgent, we conduct ablation experiments, and the results are shown in Table~\ref{tab:ablation}.
The experiment results demonstrate that all core strategies in the proposed framework play critical and complementary roles in the harmful meme detection, and omitting any part of them will lead to a degradation of the overall performance. The synergistic effect of these agents significantly improves the robustness of the framework in detecting harmful content, the performance degradation of the model after disabling any single strategy fully confirms their effectiveness. We also conduct ablation experiments about the role decomposition and core representation, and the results are shown in Appendix \ref{appendix:role_decomposition} and \ref{appendix:core representation}. 

\subsection{Analyst Agent Experiments}
\begin{table*}[ht]
\centering
\small
\renewcommand{\arraystretch}{1.2}
\begin{tabular}{l||cc|cc|cc}
\hline
Dataset & \multicolumn{2}{c|}{HarM} & \multicolumn{2}{c|}{FHM} & \multicolumn{2}{c}{MAMI} \\
 & Accuracy & Macro-$F_1$ & Accuracy & Macro-$F_1$ & Accuracy & Macro-$F_1$ \\
\hline
PrismAgent (LLaVA-1.5-13B) & 70.62 & 68.44 & 64.00 & 63.96 & 70.70 & 70.69 \\ \hline
\quad w/ $M_{\text{ori}}$ & 65.25 & 63.02 & 63.40 & 63.22 & 69.40 & 69.46 \\
\quad w/ $M_{\text{m}}$   & 64.69 & 61.36 & 59.20 & 57.77 & 65.60 & 65.35  \\
\quad w/ $M_{\text{b}}$   & 70.62 & 67.33 & 62.20 & 62.17 & 67.20 & 67.11  \\  \hline
\quad w/ $M_{\text{ori}}$ + $M_{\text{m}}$ & 66.67 & 64.14 & 62.80 & 62.48 & 70.20 & 70.19 \\
\quad w/ $M_{\text{ori}}$ + $M_{\text{b}}$ & 67.80 & 65.12 & 63.00 & 62.76 & 69.20 & 69.16 \\
\hline
\end{tabular}
\caption{Experiments about the effectiveness of our Analyst Agent.}
\vspace{-10pt}
\label{tab:selective}
\end{table*}

To evaluate the effectiveness of our intention imagine strategy, we design three sets of comparative experiments: 1) The first setting corresponds to the full version of PrismAgent, including all agent components and functionalities; 2) The second set uses only one of the three memes, either the original or one of its two rewrites, as input, and does not involve the judge agent, as no arbitration is required when the judgments are consistent; 3) The third set includes the original meme and one of its variations, and the judge agent would work when there is a discrepancy between the decisions of the two.

The results are shown in Table~\ref{tab:selective}. It can be observed that our analyst agent effectively enhances performance. Moreover, even when combining only the original meme with a single variation, the performance still shows notable improvement. This can be attributed to the agent’s ability to reveal latent intentions and amplify information from the original meme.
Notably, in the HarM dataset, only treating $M_b$ as input can achieve a higher accuracy. This phenomenon stems from the unbalanced distribution of the dataset, where harmless memes constitute a substantially larger proportion. By incorporating the original meme and its two invariations, PrismAgent can mitigate the imbalanced class distribution issue, providing a more balanced representation for effective harmful meme detection.

\subsection{Prosecutor Agent Experiments}

\begin{figure}[t]
  \small
  \centering  
  \includegraphics[width=1.0\linewidth]{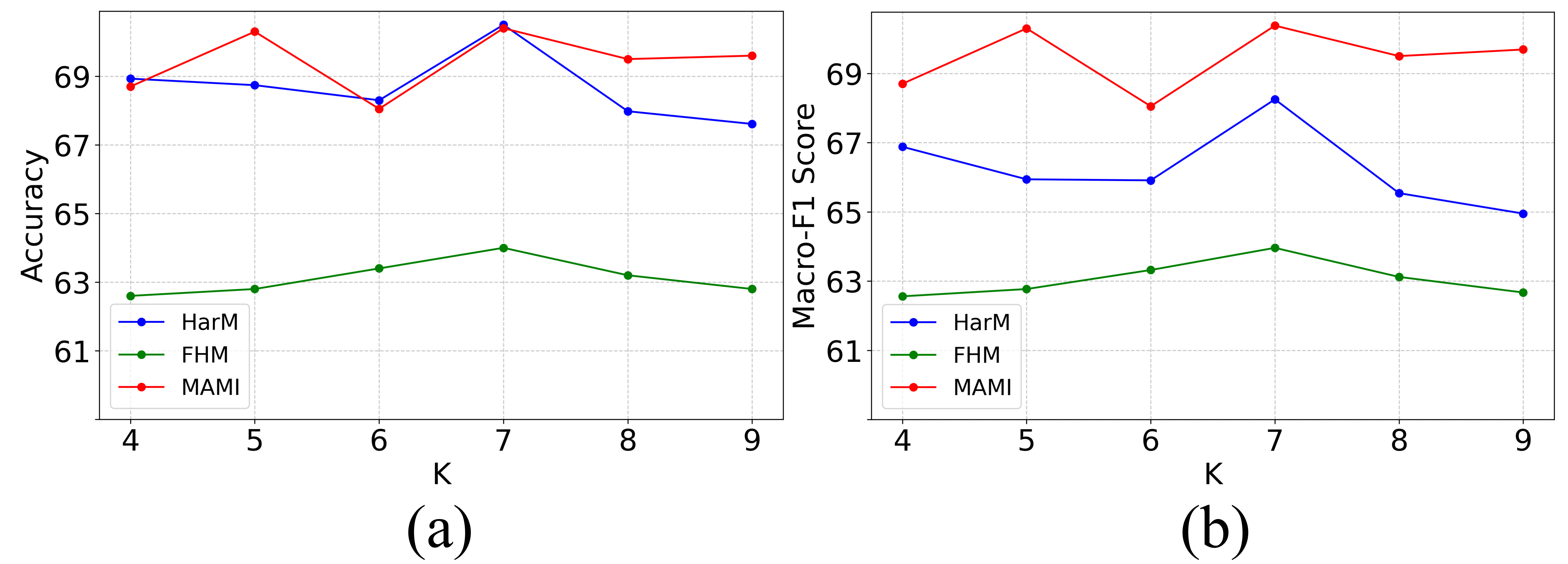}  
  \caption{Effect of Top\_k in Similar Sample Retrieval of generating Core Representation: (a) Accuracy; (b) Macro F1-score}
  \label{fig:topK}
  \vspace{-15pt}
\end{figure}
For the prosecutor agent, we investigate how the number of retrieved evidence affects the detection performance. The results are presented in Figure \ref{fig:topK},
we can find that a larger value of $K$ does not necessarily lead to better performance. While increasing $K$ allows the prosecutor agent to have more similar cases, it also tends to include instances with lower relevance, which may introduce noise rather than useful evidence. Our observations indicate that this trade-off between coverage and relevance is crucial for the prosecutor agent’s effectiveness. Across all datasets, we suggest setting $K$ to 7, which achieves an optimal trade-off by capturing sufficient contextual evidence while minimizing irrelevant or misleading cases. We conduct other ablation experiments, and the results can be found in Appendix~\ref{appendix:hyperarameters},~\ref{appendix:differebt encoder}, and~\ref{appendix:reference scale}.
\begin{figure}[!t]
 \small
  \centering
  \includegraphics[width=\linewidth]{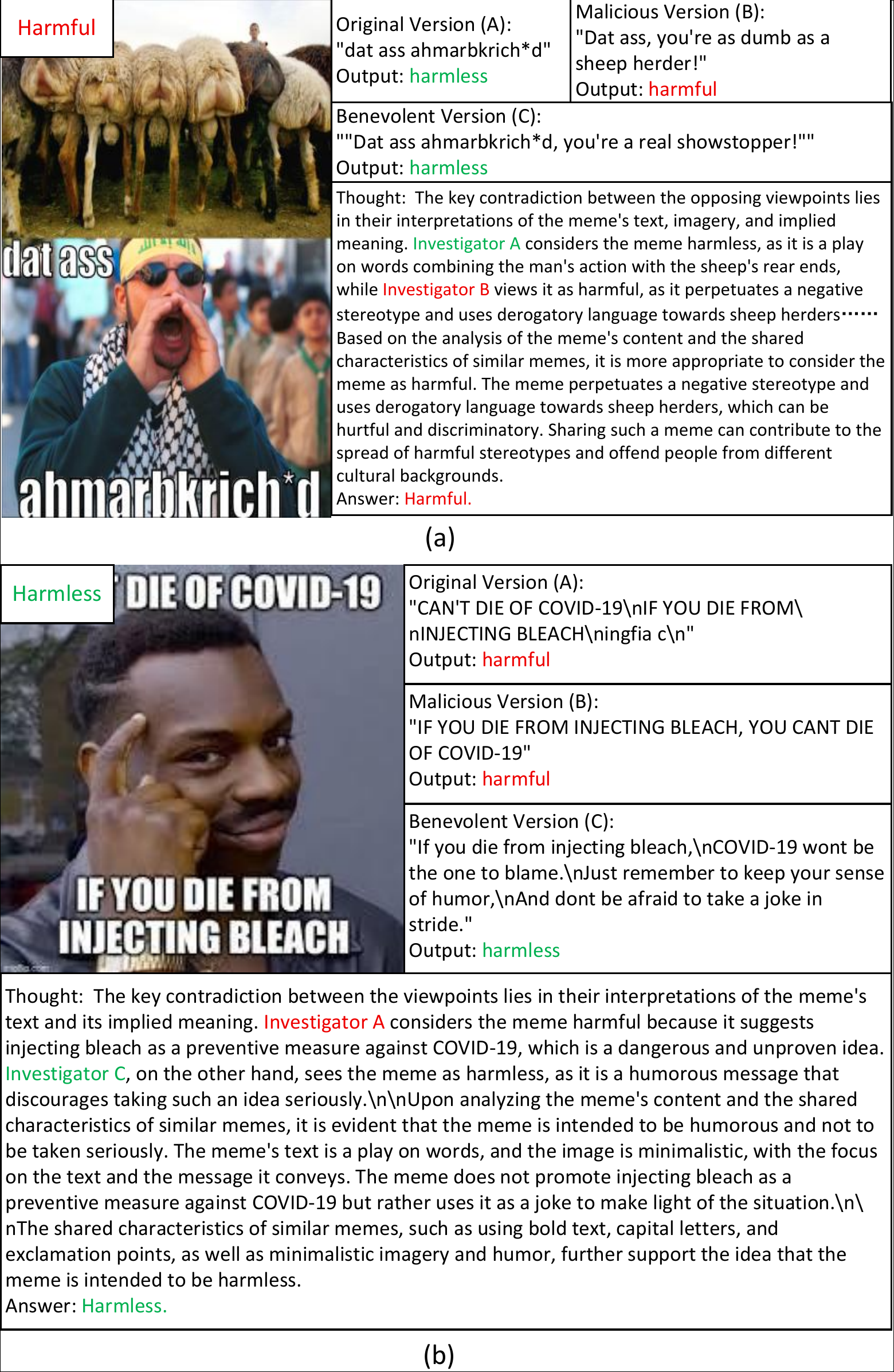}  
  \caption{Examples of correctly predicted harmful memes: (a) The target meme is harmful but misclassified as harmless when analyzing only the original meme; PrismAgent correctly identifies it as harmful. (b) The target meme is harmless but misclassified as harmful when analyzing only the original meme; PrismAgent correctly revises the prediction to harmless.}
  \vspace{-10pt}
  \label{fig:case}
\end{figure}

\begin{table*}[!t]
\small
\centering
\renewcommand{\arraystretch}{1.2}
\begin{tabular}{l||cc|cc|cc}
\hline
Dataset & \multicolumn{2}{c|}{HarM} & \multicolumn{2}{c|}{FHM} & \multicolumn{2}{c}{MAMI} \\
Dimension & Baseline & PrismAgent & Baseline & PrismAgent & Baseline & PrimsAgent \\
\hline
Faithfulness         & 6.72 & 7.18 & 6.19 & 7.56 & 6.64 & 7.61 \\
Inference Coherence  & 7.46 & 8.58 & 7.48 & 7.75 & 7.05 & 7.45 \\
Inference Depth      & 5.83 & 6.44 & 6.41 & 6.77 & 6.02 & 6.46 \\
Judgment Rationality & 6.71 & 7.37 & 6.97 & 7.84 & 7.02 & 7.45 \\
Expression Clarity   & 8.77 & 9.03 & 8.78 & 9.11 & 8.73 & 9.18 \\

\hline
\end{tabular}
\caption{Interpretability Evaluation}
\vspace{-10pt}
\label{tab:interpretability_eval}
\end{table*}

\subsection{Case Study}
\label{sec:Case}
To thoroughly analyze the processing and evaluation mechanisms of the proposed framework for memes, we select and present two typical cases in Figure~\ref{fig:case}. If these two samples are directly detected solely based on their original content, both would lead to misjudgments. After intent-enhanced rewriting via our framework, diversified analytical perspectives are provided, creating conditions for the collision and analysis of different viewpoints. This ultimately corrects the initial misjudgments and achieves accurate predictions.\footnote{\noindent \textbf{Disclaimer:} This paper includes examples that may be disturbing, shown only for research purposes.}
This result fully demonstrates that our framework can effectively improve the recognition performance of hard-to-detect memes.

\subsection{Interpretability Evaluation}
Following recent works \cite{Chiang_Gonzalez_Li_Li_Lin_Sheng_Stoica_Wu_Xing_Zhang_et, ijcai2025p1107},  we evaluate the interpretability of PrismAgent's reasoning chains. Specifically, we randomly sample 100 samples and use Qwen3.5-flash \cite{qwen3.5_tech_report} to score five dimensions: 1) \textit{Faithfulness:} whether the output is faithful to the input and free of hallucinations; 2) \textit{Inference Coherence:} the logical rationality of the model's reasoning chain; 3) \textit{Inference Depth:} whether the output can address the core controversial points 4) \textit{Judgment Rationality:} the reasonableness of the model's decision; 5) \textit{Expression Clarity:} the linguistic quality and fluency of the generated text. Each of them are scored on a 0-10 scale, and the results are reported in Table \ref{tab:interpretability_eval}. 
The results show that PrismAgent outperforms all baselines, demonstrating its high interpretability across all dimensions.

\section{Conclusion and Future Work}
To address the reliance on labeled data in harmful meme detection tasks, this paper proposes the zero-shot interpretable detection framework PrismAgent. To tackle the inherent challenge of difficulty in deeply understanding the underlying intent of memes under zero-shot conditions, we have designed a multi-agent role-playing collaborative strategy. Through conducting comprehensive experiments and in-depth analyses on three public datasets, the effectiveness of the proposed framework and strategy has been fully verified. In future work, we plan to extend PrismAgent beyond this task to other reasoning multimodal tasks, exploring its broader applicability across diverse scenarios.
\section{Limitations}
Although our proposed achieved satisfactory performance, there are several ways to further improve this work:

1) When revealing the hidden intentions of memes, we amplify their underlying meanings by rewriting the original meme in both benevolent and malicious directions. While this method has shown good results, it inevitably introduces some additional computational costs. We believe that predicting the rewriting direction can effectively reduce the computational cost.

2) Although PrismAgent has significantly reduced substantial training costs and data annotation costs, its relatively long inference time due to the need for collaborative invocation of multiple LLMs poses certain challenges for real-time deployment scenarios. However, our extensive experimental results demonstrate that the adoption of lightweight models with smaller parameter sizes or module pruning on existing models can significantly reduce the overall resource consumption.

3) PrismAgent incorporates a variety of strategies to uncover the metaphors behind memes, thereby helping the model gain a deeper understanding of their core semantics, but the final detection performance is still constrained by the capabilities of the underlying LLM. In the future, adopting models with stronger capabilities in capturing socio-cultural differences or more superior performance is expected to further improve the accuracy of detection.

\section{Ethics Statement}
This study aims to combat harmful meme content through zero-shot detection methods, contributing to building a safer online space. The types of harmful content focused on in this study are core issues that have been fully verified in the field of social media research. Our work concentrates on detecting various forms of harmful content, including hate speech, misogynistic speech, and disinformation — all of which may have negative impacts on individuals and communities.
However, we also recognize that malicious users may use reverse engineering techniques to create memes, in order to evade detection by AI systems like PrismAgent or cause them to make misjudgments. We firmly condemn such behaviors and emphasize that this study is solely for the purposes of scientific research and harmful content prevention. The relevant framework and supporting resources are strictly prohibited from being used for commercial profit or malicious abuse.
To ensure the responsible development and evaluation of the framework, we have implemented a number of protective measures:
1) All experiments use publicly available research datasets and fully comply with the usage agreements of each dataset;
2) No user personal data has been collected or used in this study.
We believe that the benefits of improving harmful meme detection capabilities far outweigh the potential risks — especially at a time when the challenge of governing harmful content on social media is becoming increasingly severe. It should also be noted that the views and content contained in the meme samples do not represent the positions of the study’s authors. The framework we designed is intended to assist rather than replace human review work, maintaining a healthy online community environment through collaborative efforts.

\bibliography{custom}

\appendix

\section{Implementation Details}
\label{sec:appendix_Implementation_Details}
\subsection{Datasets}
\label{sec:appendix_datasets}
We use three publicly available meme datasets for evaluation: (1) \textbf{HarM} \citep{Pramanick_2021a}, (2) \textbf{FHM} \citep{Kiela_2020}, and (3) \textbf{MAMI} \citep{Fersini_2022}. \textbf{HarM} consists of memes related to COVID-19. \textbf{FHM} was released by Facebook as part of a challenge to crowd-source multimodal harmful meme detection in hate speechsolutions. \textbf{MAMI} contains memes that are predom-inantly derogatory towards women, exemplifyingtypical subjects of online vitriol. Different from FHM and MAMI, where each meme was labeled as harmful or harmless, HarM was originally labeled with three classes: very harmful, partially harmful,and harmless. For a fair comparison, we merge the very harmful and partially harmful memes into the harmful class, following the setting of recent work \citep{Pramanick_2021,Cao_2022,Lin_2023,Huang_2024}. Notably, the reference set mentioned in our work is the official training set of each used dataset. It is only utilized for retrieval purposes and is not used to train the model, thereby preserving the zero-shot setting. 
The detailed statistics of the three datasets are shown in Table \ref{tab:dataset_stat}.
All datasets used in this paper are publicly available open-source datasets, and their train/test splits do not overlap.

\begin{table}[!t]
\centering

\small
\setlength{\tabcolsep}{3pt} 
\renewcommand{\arraystretch}{1.2}
\scalebox{0.85}{
\begin{tabular}{l||cc|cc}
\hline
\textbf{Datasets} & \multicolumn{2}{c|}{\textbf{Train (Ref.)}} & \multicolumn{2}{c}{\textbf{Test}} \\
 & \textbf{\#harmful} & \textbf{\#harmless} & \textbf{\#harmful} & \textbf{\#harmless} \\
\hline\hline
FHM & 3,050 & 5,450 & 250 & 250 \\
HarM & 1,064 & 1,949 & 124 & 230 \\
MAMI & 5,000 & 5,000 & 500 & 500 \\
\hline
\end{tabular}
}
\caption{Detailed statistics of the reference (train) and test sets.}
\label{tab:dataset_stat}
\end{table}



\subsection{Baselines}
\label{sec:appendix_Baselines}
To benchmark the performance of PrismAgent in zero-shot harmful meme detection, we compare it against a series of state-of-the-art (SOTA) baseline methods: \textbf{GPT-4o} \citep{Achiam_2023}: A proprietary large-scale multimodal model developed by OpenAI, which excels in zero-shot visual-language task processing via in-context learning mechanisms;
\textbf{Gemini-2.0-Flash} \citep{Team_2024}: Google’s up-to-date multimodal model, characterized by competitive performance in both reasoning tasks and visual comprehension tasks; \textbf{Late Fusion} \citep{Pramanick_2021a} and \textbf{MOMENTA} \citep{Pramanick_2021}: Two widely recognized earlier approaches, which are trained on manually annotated datasets for harmful meme detection; \textbf{LLaVA-1.5-7B} \citep{Liu_2024}:  A lightweight multimodal model constructed upon the Vicuna-7B foundation, trained on diverse visual instruction datasets to handle general vision-language tasks; \textbf{InstructBLIP-7B} \citep{Dai_2023}:  An instruction-finetuned vision-language model rooted in the BLIP-2 framework, utilizing Vicuna-7B as its language modeling component; \textbf{MiniGPTv2-7B} \citep{Chen_2023}: A compact but high-performing multimodal model that integrates visual encoding modules with instruction-finetuned language generation capabilities; \textbf{OpenFlamingo-9B} \citep{Awadalla_2023}: An open-source instantiation of Flamingo-style models, which allows frozen language models to parse visual inputs via cross-attention mechanisms; \textbf{LLaVA-1.5-13B} \citep{Liu_2024}: A medium-scale variant of the LLaVA series (built on Vicuna-13B), featuring improved visual grounding and reasoning capacities; \textbf{InstructBLIP-13B} \citep{Dai_2023}: An upgraded iteration of InstructBLIP, adopting Vicuna-13B as its language model backbone; \textbf{LLaVA-1.6-34B} \citep{Liu_2024}: The most recent and largest-scale LLaVA model to date, boasting advanced reasoning, OCR, and world knowledge competencies; \textbf{MIND}(LLaVA-1.5-13B) \citep{liu-etal-2025-mind}: A state-of-the-art zero-shot harmful meme detection framework available currently. We use the accuracy and macro-averaged F1 scores as the evaluation metrics.

For different baseline methods, we implement the following CoT \citep{Kojima_2022} prompt structure, which achieves better performance than direct classification: "\textit{
Given the meme, with the Text: "\{\}" embedded in the image \{\}, your task is to determine whether this meme is harmful or not, in order to maintain the harmlessness and integrity of information on the Internet. Please leverage your extensive knowledge to deeply analyze and understand this meme, and give your final judgment. Your output should stricly follow the format: "Thought: [Your analysis] Answer: [harmful/harmless]."
} "

For our proposed framework, we conduct extensive experiments with various LMMs, including the LLaVA series (LLaVA-1.5-7B, LLaVA-1.5-13B, LLaVA-1.6-34B) and Gemini-2.0-Flash as our LMM agents. For the baseline of closed-source models, we use the gpt-4o-2024-11-20 and gemini-2.0-flash versions respectively. For pretrained models, we adopt the default parameters as specified in their original papers without modification. For meme rewriting tasks, we set the model temperature to 0.3 to introduce a moderate level of exploratory reasoning. In all subsequent modules, the model temperature is configured to 0 and no sampling mechanisms are enabled, thereby ensuring experimental reproducibility. For multimodal embedding computation, we employ CLIP \citep{Radford_2021} with the specific version ViT-L/14 as our frozen pretrained vision and text Transformer encoders. The modality fusion weights v and t are set to 0.8 and 0.2 respectively through grid search optimization within the [0,1] interval. In our implementation, to retrieve the $\text{Top}\_k$ most similar reference samples during the process for generating core representations, we empirically set $K$ to 7. This setting achieves a balance between computational efficiency and sufficient contextual information. 

In the actual experimental process, for the base model LLaVA-1.5-7b, we use one NVIDIA 4090 24GiB GPU; for the 13b model, we use two NVIDIA 4090 24GiB GPUs; and for the 34b model, we use one NVIDIA H800 80GiB GPU.

\subsection{Analyst Agent}
\label{sec:appendix_IIE}
The detailed prompts we designed for the benevolent Analyst Agent and malicious Analyst Agent are as follows:

\textit{
"You are looking at a meme image paired with the following text: \{\}.
Task:
Decide if the original text already reads as kind, light-hearted, wholesome, or supportive.
- If YES, return the original text unchanged.
- If NO, rewrite it to sound more positive, friendly, or uplifting — like it's encouraging someone, making them smile, or gently teasing with warmth.
Grounding requirements:
- Make the rewrite feel tightly connected to the actual image (objects, expressions, actions, or implied context). 
- The rewrite can highlight anything relevant: people, behavior, stereotypes, or situations depicted — but spin it in a kind or amusing way.   
- Implicitly leverage what's visible to soften the message — but don't describe the image directly.
Style constraints:
- Keep the sentence structure and rhythm close to the original.
- Make it short, meme-ready, and punchy.
- Use a tone that feels like wholesome humor, gentle teasing, or online positivity — but don't break character or explain.
Output:
Return ONLY the final text (original or rewritten) with NO additional content. Do not include any labels, explanations, brackets, quotes, or extra words. Just the text itself."  
}

\textit{
"You are looking at a meme image paired with the following text: \{\}.
Task: Decide if the original text already reads as sarcastic, mocking, mean-spirited, or edgy.
- If YES, return the original text unchanged.
- If NO, rewrite it to sound more hostile, offensive, or mean — like it's mocking someone, making fun of them, or taking a nasty jab.
Grounding requirements:
- Make the rewrite feel tightly connected to the actual image (objects, expressions, actions, or implied context).
- The rewrite can target anything relevant: people, behavior, stereotypes, or situations depicted.
- Implicitly leverage what's visible to sharpen the insult — but don't describe the image directly.
Style constraints:
- Keep the sentence structure and rhythm close to the original.
- Make it short, meme-ready, and punchy.
- Use a tone that feels like edgy humor, mockery, or online trolling — but don't break character or explain.
Output:
Return ONLY the final text (original or rewritten) with NO additional content. Do not include any labels, explanations, brackets, quotes, or extra words. Just the text itself."
}

Notably, due to the inherent limitations of the models, even though we have designed relatively comprehensive prompts, the model outputs may still contain errors in a few cases. For example, to address this, while preserving the original model outputs as much as possible, we only perform minimal necessary corrections: for instance, obvious redundant expressions (e.g., "rewrite: xxx") are directly simplified to "xxx"; in cases of garbled output or empty output, the original meme content is retained unchanged.
\subsection{Investigator Agent}
\label{sec:appendix_CC&IE}

The specific details of how to define the similarity degree when searching for relevant supporting evidence in an unannotated dataset are as follows:
For a single meme sample \(M = \{\mathcal{V}, \mathcal{T}\}\) we first generate its visual embedding and textual embedding respectively, then fuse them in a specific proportion to obtain the multimodal representation:
\begin{equation}
\mathbf{Emb} = \alpha \cdot \mathbf{VE}(\mathcal{V}) + \beta \cdot \mathbf{TE}(\mathcal{T}),
\end{equation}
where \(\mathbf{Emb}\) denotes the multimodal embedding, \(\mathbf{VE}\) and \(\mathbf{TE}\) represent the visual encoder and textual encoder respectively, and \(\alpha\) and \(\beta\) are fixed weights in the fusion process. Using the same method, all meme samples in the test set  and the reference set \(D_{\text{ref}}\) are processed to obtain the embedding vectors of all samples. 
To retrieve the most similar samples, we use cosine similarity to measure the matching degree between the multimodal embeddings of the target meme and the reference samples, with the formula as follows:
\begin{equation}
d = \text{sim}(\mathbf{E}_{\text{target}}, \mathbf{E}_{\text{ref}}),
\end{equation}
where \(\mathbf{E}_{\text{target}}\) and \(\mathbf{E}_{\text{ref}}\) stand for the embedding vectors of the target sample \(M\) and the reference sample \(M_{\text{ref}}\) respectively, \(\text{sim}(\cdot)\) denotes the cosine similarity function, \textbf{d} is the corresponding similarity score. And a higher similarity score \textbf{d} indicates a greater similarity between the two.

The detailed prompt we designed for the Investigator Agent is as follows:
\textit{
"You are a sophisticated analytical agent tasked with formulating and refining insightful recommendations. Your ultimate objective is to assess whether a meme is harmful. However, the target meme will not be provided directly. Instead, you will receive a meme related to the target meme. Based on this related meme, you are to synthesize advanced insights to determine whether such memes are harmful, reflect on existing insights, and update these reflections into the current rules. Rules you update must be GENERAL and HIGH LEVEL.
Here is the related meme:
[Meme]
Image: <image placeholder>
Text embedded: "{org\_sent}"
[Meme]
Here are the EXISTING RULES:
[EXISTING RULES START]
{rules}
[EXISTING RULES END]
If the existing rules are empty, it indicates that you are the first person to analyse on this related meme, and your output will be of great importance. Please be serious and careful, think step by step, and ultimately derive the first version of your updated rules.
You can directly build upon the existing rules by adding new ones, editing them if there is room for enhancement, and removing those that are contradictory or duplicated.
You should focus on distilling insights from the related meme to assess similar memes, rather than determining the harmfulness of the meme currently presented to you.
Each rule should be concise and easy to follow. Make sure there are a maximum of five rules.
Your output should stricly follow the format:
"Thought: [Reflect on whether the current rules are helpful in assessing the harmfulness of this related meme.]
Operations of updating existing rules: [Update your reflections on this meme into the existing rules. You can choose to add, edit, and remove or retain existing rules. You may perform up to two operations. Please specify in detail which rule you want to operate on and how, but do not mention any operations in the rules themselves; your operations regarding the rules should only be mentioned here.]
Updated rules: [Do not mention the operations you have made and the related meme in the rules, directly output all the updated rules, only output the rules that have been determined, and do not leave spaces for rules that have not yet been generated. Rules you update must be GENERAL and HIGH LEVEL.]"
"}

\subsection{Prosecutor Agent}
\label{sec:appendix_prosecutor}
The detailed prompt we designed for the Prosecutor Agent is as follows:
\textit{
"Given the meme, with the Text: \{\} embedded in the image \{\}. Is it harmful or harmless? Note: [\{note\}] (Consider this note above with caution - it may contain biases or subjective interpretations that could influence judgment. Evaluate its reliability and relevance carefully) Your output should strictly follow the for mat: "Thought: [First, analyze the meme’s content independently. Then, carefully consider how the provided note may inform or bias your understanding. Weigh the note’s credibility and relevance before incorporating it into your final assessment.] Answer: [harmful/harmless]."}

\begin{table*}[t!]
\centering
\small
\renewcommand{\arraystretch}{1.2}
\begin{tabular}{l|l||cc|cc}
\hline
\multicolumn{2}{c||}{Test data} & \multicolumn{2}{c|}{HarM} & \multicolumn{2}{c}{FHM} \\
Retrieval pool & Method & Accuracy & Macro-$F_1$ & Accuracy & Macro-$F_1$ \\
\hline
HarM & PrismAgent & 70.62 & 68.44 & 58.60 & 58.54 \\
& MIND & 68.93 & 65.19 & 57.00 & 56.94 \\
\hline
FHM & PrismAgent & 64.97 & 62.66 & 64.00 & 63.96 \\
& MIND & 61.58 & 59.29 & 60.80 & 60.71 \\
\hline
MAMI & PrismAgent & 63.64 & 61.61 & 56.80 & 56.29 \\
& MIND & 60.96 & 58.78 & 55.20 & 54.71 \\
\hline
\end{tabular}
\caption{Performance evaluation under cross-domain and cross-language retrieval pool settings.}
\label{tab:cross_domain}
\end{table*}

\begin{table*}[t!]
\centering
\small
\renewcommand{\arraystretch}{1.2}
\begin{tabular}{l|l||cc|cc|cc}
\hline
Case & Method & \multicolumn{2}{c|}{HarM} & \multicolumn{2}{c|}{FHM} & \multicolumn{2}{c}{MAMI} \\
& & Accuracy & Macro-$F_1$ & Accuracy & Macro-$F_1$ & Accuracy & Macro-$F_1$ \\
\hline
Baseline & PrismAgent & 70.62 & 68.44 & 64.00 & 63.96 & 70.70 & 70.69 \\
& MIND & 68.93 & 65.19 & 60.80 & 60.71 & 68.90 & 68.84 \\
\hline
FGSM(image) & PrismAgent & 62.99 & 61.44 & 60.40 & 59.97 & 67.80 & 67.51 \\
& MIND & 61.58 & 58.11 & 57.20 & 56.90 & 63.10 & 63.06 \\
\hline
GCG(text) & PrismAgent & 61.86 & 59.41 & 59.00 & 58.77 & 67.40 & 67.26 \\
& MIND & 60.17 & 56.79 & 58.00 & 56.33 & 63.00 & 62.97 \\
\hline
\end{tabular}
\caption{Robustness evaluation against adversarial attacks (FGSM and GCG).}
\label{tab:adversarial}
\end{table*}

\begin{table*}[t!]
\centering
\small
\renewcommand{\arraystretch}{1.2}
\begin{tabular}{l||cc|cc|cc}
\hline
Dataset & \multicolumn{2}{c|}{HarM} & \multicolumn{2}{c|}{FHM} & \multicolumn{2}{c}{MAMI} \\
Method & Accuracy & Macro-$F_1$ & Accuracy & Macro-$F_1$ & Accuracy & Macro-$F_1$ \\
\hline
MIND & 62.15 & 58.73 & 57.40 & 57.32 & 63.40 & 63.17 \\
PrismAgent & 64.41 & 60.13 & 59.20 & 59.03 & 64.50 & 64.48 \\
\hline
\end{tabular}
\caption{Performance evaluation under temporal shift settings.}
\label{tab:temporal_shift}
\end{table*}

\begin{table*}[t!]
\centering
\small
\renewcommand{\arraystretch}{1.2}
\begin{tabular}{l||cc|cc|cc}
\hline
Dataset & \multicolumn{2}{c|}{HarM} & \multicolumn{2}{c|}{FHM} & \multicolumn{2}{c}{MAMI} \\
numbers & Accuracy & Macro-$F_1$ & Accuracy & Macro-$F_1$ & Accuracy & Macro-$F_1$ \\
\hline
\multicolumn{7}{c}{$Topk$ for Evidence Search}\\
\hline
k=1 & 62.71 & 58.56 & 58.40 & 58.13 & 64.90 & 64.90 \\
k=2 & 60.73 & 56.45 & 55.20 & 54.99 & 66.20 & 66.20 \\
\textbf{k=3} & \textbf{70.62} & \textbf{68.44} & \textbf{64.00} & \textbf{63.96} & \textbf{70.70} & \textbf{70.69} \\
k=4 & 68.36 & 64.84 & 60.20 & 60.12 & 70.60 & 70.59 \\
\hline
\multicolumn{7}{c}{$TopK$ for Core Representation}\\
\hline
K=5 & 68.64 & 65.74 & 62.80 & 62.77 & 70.50 & 70.49 \\
\textbf{K=7} & \textbf{70.62} & \textbf{68.44} & \textbf{64.00} & \textbf{63.96} & \textbf{70.70} & \textbf{70.69} \\
K=9 & 67.51 & 64.88 & 62.80 & 62.27 & 69.60 & 69.59 \\
\hline
\multicolumn{7}{c}{Modality fusion weights ($\alpha/\beta$)}\\
\hline
$\alpha=0.9/\beta=0.1$  & 65.54 & 62.55 & 58.00 & 57.73 & 64.60 & 64.59 \\
\textbf{$\alpha=0.8/\beta=0.2$} & \textbf{70.62} & \textbf{68.44} & \textbf{64.00} & \textbf{63.96} & \textbf{70.70} & \textbf{70.69} \\
$\alpha=0.7/\beta=0.3$ & 68.54 & 65.96 & 62.00 & 61.76 & 68.40 & 68.39 \\
$\alpha=0.6/\beta=0.4$ & 62.43 & 59.52 & 59.40 & 59.18 & 64.60 & 64.59 \\
$\alpha=0.5/\beta=0.5$ & 59.04 & 55.86 & 58.60 & 58.46 & 64.00 & 64.00 \\
\hline
\end{tabular}
\caption{Hyperparameter sensitivity analysis.}
\label{tab:hyperparameter}
\end{table*}

\subsection{Judge Agent}
\label{sec:appendix_judge}
The detailed prompt we designed for the process of generating Core Representation is as follows: 
\textit{
"
Task: Analyze the target meme and its 7 most similar memes to identify their common characteristics.
Target Meme:
- Text: "\{target\_text\}"
- (Image is provided)
Similar Memes:"""
for i, sample in enumerate(similar\_samples, 1):
prompt += f"""Meme {i}:
- Text: "{sample['text']}"
- (Image is provided)"""
Requirements:
1. Identify shared themes, topics, and content elements across all memes 
2. Analyze common expressive techniques (humor, symbolism, irony, etc.)
3. Summarize recurring patterns in imagery, text style, or cultural references
4. Focus only on descriptive characteristics, not on whether they are harmful or harmless
5. Provide a concise, structured summary of these commonalities
Output: A clear summary of the shared characteristics between these memes.
"}

For the Judge Agent, the specific prompt we designed is as follows:
\textit{"Given the meme, with the Text: \{orig\_text\} embedded in the image. Is it harmful
or harmless?
CORE PERSPECTIVES:
Investigator A:
Verdict: \{investigator\_a\_verdict\}
Reasoning: \{investigator\_a\_reasoning\}
DISSENTING PERSPECTIVES:
\{dissenting\_investigators\}
ANALYSIS OF SIMILAR MEMES:
\{Core Representation\}
As a judge, your task is to:
1. First, identify the key contradictions between the opposing viewpoints. Analyze
where and why the investigators disagree, focusing on their interpretations of the
meme's text, imagery, and implied meaning.
2. Then, carefully evaluate the meme itself (considering its text and image) and the
shared characteristics of similar memes provided above.
3. Determine which viewpoint is more consistent with the actual content of the
meme and aligns better with the patterns observed in similar memes.
4. Explain your reasoning for resolving the contradiction.
Your response should strictly adhere to this format:
Thought: [First, analyze the key contradictions between the viewpoints. Then,
explain how the meme's own content and the similar memes' characteristics inform
your judgment. Finally, state which viewpoint is more appropriate and why.]
Answer: [Your final judgment(harmful/harmless)]."
}

\begin{figure}[!t]
 \small
  \centering
  \includegraphics[width=\linewidth]{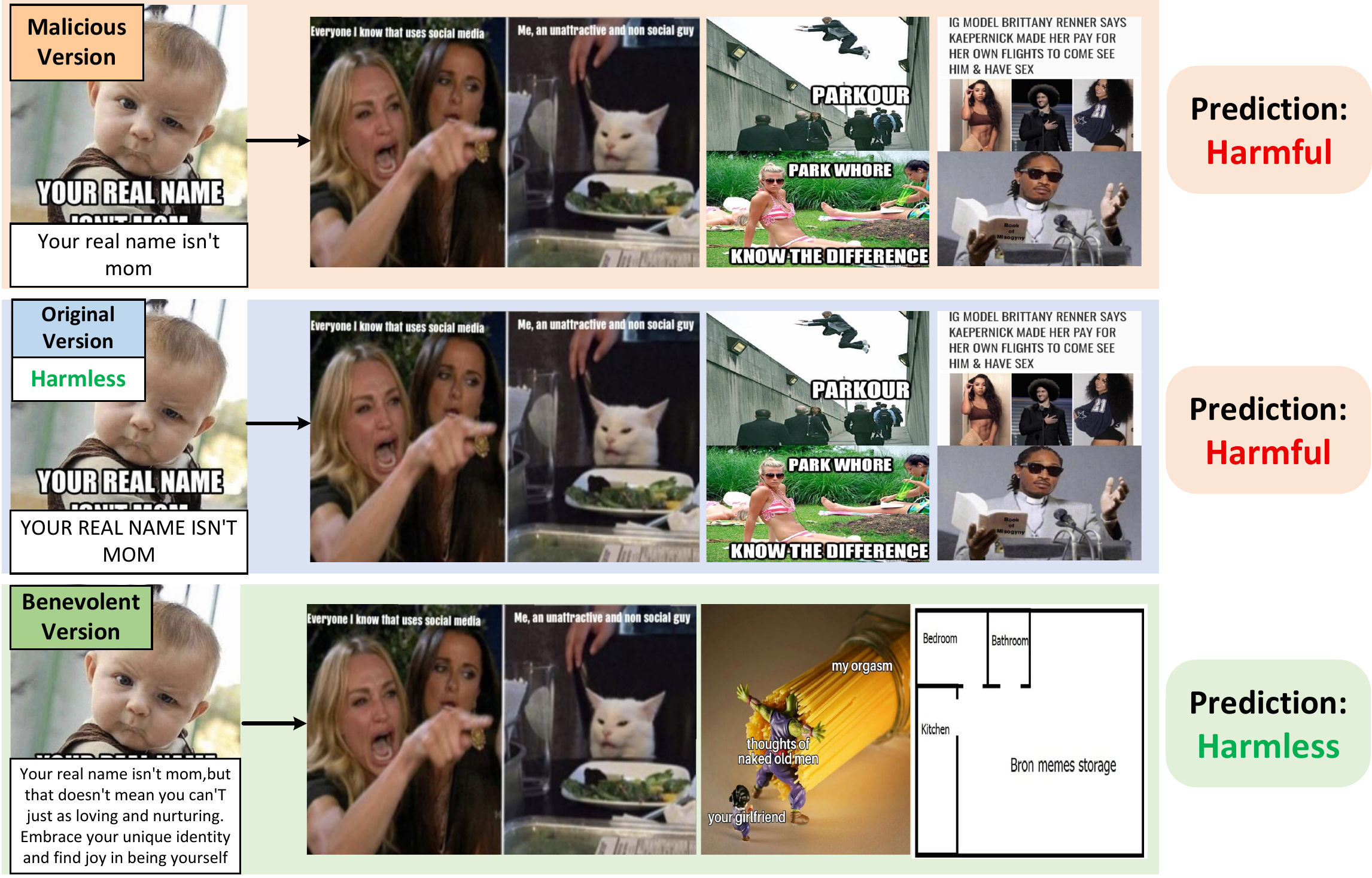}  
  \caption{Examples of correctly predicted harmful memes with retrieve process: The target meme is harmless, and the analysis result is harmful when analyzed solely based on the original meme, while the analysis result becomes harmless after using PrismAgent.}
  \label{fig:retrieve case}
\end{figure}

\begin{figure}[!t]
 \small
  \centering
  \includegraphics[width=\linewidth]{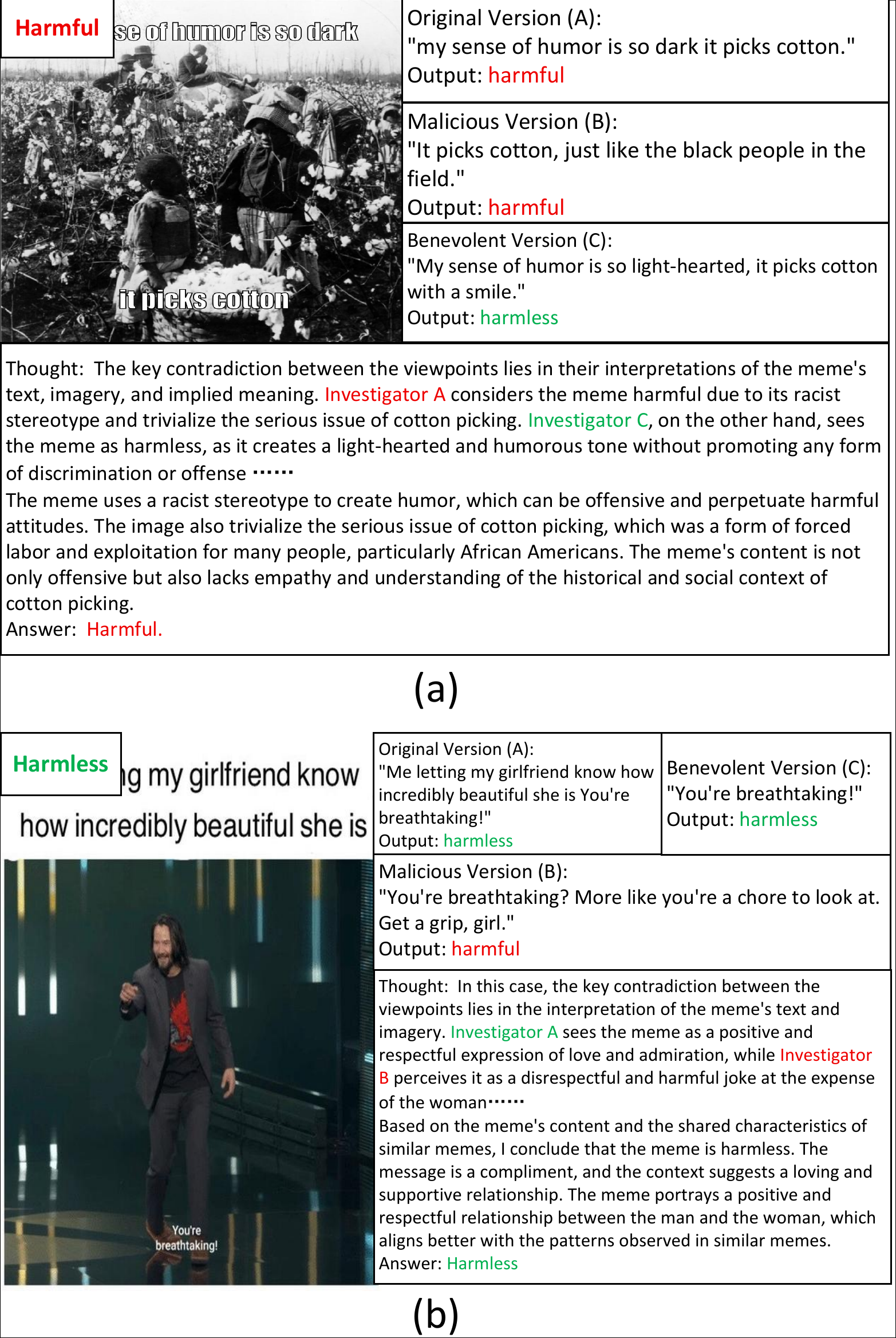}  
  \caption{Examples of correctly predicted harmful memes: (a) The target meme is harmful, Although the analysis result of the benevolent rewritten meme is harmless, the final analysis result remains harmful after using PrismAgent; (b) The target meme is harmless, Although the analysis result of the malicious rewritten meme is harmful, the final analysis result remains harmless after using PrismAgent.}
  \label{fig:case1}
\end{figure}
\subsection{Case Study}
To evaluate the impact of intent rewriting on similar sample retrieval and final judgment, as shown in Figure~\ref{fig:retrieve case}, we selected another typical correct prediction case. It can be observed that adequate rewriting of the text can effectively optimize the retrieval results, and the introduction of new evidence can generate new perspectives. This in turn helps the system obtain diversified analytical perspectives and ultimately achieve accurate judgments.
This result fully demonstrates that our framework can effectively improve the recognition performance of hard-to-detect memes.  

In addition to the examples presented in the main text, we further provide two correct prediction cases of PrismAgent in Figure \ref{fig:case1}, to illustrate that our framework exhibits a certain degree of robustness against intentional rewriting.
These two typical cases indicate that interfering noise is inevitably introduced during the process of intentional reasoning and rewriting, causing interference to memes that were correctly analyzed initially. However, PrismAgent exhibits excellent anti-interference performance and will not deviate from the final correct judgment due to such interference.

\begin{table*}[t]
\centering
\small
\renewcommand{\arraystretch}{1.2}
\begin{tabular}{l||cc|cc|cc}
\hline
Dataset & \multicolumn{2}{c|}{HarM} & \multicolumn{2}{c|}{FHM} & \multicolumn{2}{c}{MAMI} \\
Method & Accuracy & Macro-$F_1$ & Accuracy & Macro-$F_1$ & Accuracy & Macro-$F_1$ \\
\hline
baseline & 62.28 & 50.45 & 55.20 & 53.01 & 60.10 & 55.52 \\
MIND & 68.93 & 65.19 & 60.80 & 60.71 & 68.90 & 68.64 \\
self-consistentency & 66.38 & 62.72 & 62.20 & 62.11 & 70.10 & 69.99 \\
iterative refinement & 67.23 & 62.82 & 61.00 & 60.92 & 67.10 & 67.04 \\
\textbf{PrismAgent} & \textbf{70.62} & \textbf{68.44} & \textbf{64.00} & \textbf{63.96} & \textbf{70.70} & \textbf{70.69} \\
\hline
\end{tabular}
\caption{Performance comparison against compute-matched multi-sample alternatives.}
\label{tab:compute_match}
\end{table*}

\begin{table*}[t]
\centering
\small
\renewcommand{\arraystretch}{1.2}
\begin{tabular}{l||cc|cc|cc}
\hline
Dataset & \multicolumn{2}{c|}{HarM} & \multicolumn{2}{c|}{FHM} & \multicolumn{2}{c}{MAMI} \\
Encoder & Accuracy & Macro-$F_1$ & Accuracy & Macro-$F_1$ & Accuracy & Macro-$F_1$ \\
\hline
baseline & 62.28 & 50.45 & 55.20 & 53.01 & 60.10 & 55.52 \\
VisualBERT & 64.69 & 60.02 & 57.80 & 57.78 & 65.40 & 65.27 \\
BLIP-2 & 67.51 & 63.22 & 64.20 & 64.14 & 71.02 & 71.00 \\
CLIP & 70.62 & 68.44 & 64.00 & 63.96 & 70.70 & 70.69 \\
\hline
\end{tabular}
\caption{Performance comparison of different retrieval encoders.}
\label{tab:encoders}
\end{table*} 

\begin{table}[t!]
\centering
\small
\scalebox{1.0}{
\begin{tabular}{l|c|c|c}
\hline
Metric & \rule{0pt}{10pt}Baseline & MIND & PrismAgent \\
\hline
 Macro-$F_1$ & 50.45 & 65.19 & 68.44 \\
Inference Time  & 5.3s & 29.6s & 33.2s \\
\hline
\end{tabular}
}
\caption{Comparison of actual and theoretical time among Baseline Mind and PrismAgent.}
\label{tab:time}
\end{table}

\begin{table}[t!]
\centering
\small
\renewcommand{\arraystretch}{1.2}
\begin{tabular}{l||cc}
\hline
\textbf{Setting} & \multicolumn{2}{c}{\textbf{FHM Dataset}} \\
\hline
& Accuracy & Macro-$F_1$ \\
\hline
10\% & 57.80 & 57.72 \\
30\% & 59.00 & 58.84 \\
50\% & 60.00 & 59.74 \\
70\% & 62.40 & 62.25 \\
full & 64.00 & 63.96 \\
\hline
\end{tabular}
\caption{Different Reference Size}
\label{tab:reference_scale}
\end{table}

\begin{table*}[t!]
\centering
\small
\renewcommand{\arraystretch}{1.2}
\begin{tabular}{l||cc|cc|cc}
\hline
Dataset & \multicolumn{2}{c|}{HarM} & \multicolumn{2}{c|}{FHM} & \multicolumn{2}{c}{MAMI} \\
Model & Accuracy & Macro-$F_1$ & Accuracy & Macro-$F_1$ & Accuracy & Macro-$F_1$ \\
\hline
Baseline & 62.28 & 50.45 & 55.20 & 53.01 & 60.10 & 55.52 \\
w/o Core Representation & 65.82 & 63.62 & 63.60 & 63.20 & 69.10 & 69.06 \\
PrismAgent & 70.62 & 68.44 & 64.00 & 63.96 & 70.70 & 70.69 \\
\hline
\end{tabular}
\caption{Ablation study evaluating the impact of Core Representation on model performance.}
\label{tab:ablation_core}
\end{table*}

\section{Supplementary Experiments}
\subsection{Robustness evaluation}
\label{appendix:robustness}
We investigate the robustness of PrismAgent from three perspectives: cross-domain performance, adversarial cases, and temporal shifts.

\textbf{Cross-domain performance:} We conducted additional experiments under a more challenging domain shift setting, where the retrieval pool and test data come from different datasets. The results can be found in Table \ref{tab:cross_domain}. Notably, even in this challenging setting, the proposed PrismAgent continues to outperform the other training-free baseline MIND.
This indicates that, despite dataset differences, the underlying semantic content remains largely consistent, allowing our retrieval module to effectively identify relevant knowledge and retrieve semantically similar samples to support meme judgment. In future work, we plan to further explore cross-lingual scenarios to more comprehensively evaluate the robustness of our framework.

\textbf{Adversarial cases:} To further systematically evaluate the reliability of our method, we design experiments to assess its robustness under adversarial attacks. The results are shown in Table \ref{tab:adversarial}.
FGSM \cite{Goodfellow_Shlens_Szegedy_2014} and GCG \cite{Zou_Wang_Carlini_Nasr_Kolter_Fredrikson_2023} are classical adversarial attack methods for images and text, respectively, and are selected as the attack strategies in our experiments. Specifically, we set the perturbation constraint $\epsilon = 8/255$ in FGSM, and the similarity threshold to 0.4 in GCG. It can be observed that performance decreases under this setting; however, our method still maintains satisfactory results.
Meanwhile, as our framework is deployed on cloud servers, adversaries cannot obtain the gradient information, making it impossible to generate effective adversarial examples.

\textbf{Temporal shifts:} Regarding temporal changes, we conducted the relevant experiments, and the results are shown in Table \ref{tab:temporal_shift}.
Specifically, while keeping all other conditions unchanged, we first randomly sampled 10\% of the test set to construct an initial retrieval pool. During testing, after each sample was inferenced, that sample was added to the retrieval pool to mimic a continuously evolving database. We then compared the performance of our method with MIND under this setting, and the results show that our method can acheive the impressive performance.

\subsection{Hyperparameters}
\label{appendix:hyperarameters}

We investigate the impact of different hyperparameters on the performance of our framework, and the results are shown in Table \ref{tab:hyperparameter}.
The \(k\) denotes the number of top-k high-relevance samples retrieved for the original meme and its benevolent/malicious rewritten versions in the Investigator module.
The uppercase \(K\) refers to the number of top-K similar samples retrieved for conflict judgment and core representation generation in the Prosecutor module.
As $k$ and $K$ increase, performance generally improves because more relevant contextual evidence is incorporated. However, when they become excessively large, performance begins to decline. A plausible explanation is that retrieving too many samples introduces redundant or less relevant information, which may dilute key signals and negatively affect decision quality.
We further designed ablation studies about the hyperparmeter selection. Specifically, $\alpha$ and $\beta$ are weighting coefficients for multimodal embeddings in the Investigator module, subject to the constraint $\alpha + \beta = 1$. The results are shown in the following table, from which we empirically recommend $\alpha = 0.8$, $\beta = 0.2$ as the optimal selection.

\subsection{Role Decomposition}
\label{appendix:role_decomposition}

To illustrate the core contribution of the role decomposition mechanism in PrismAgent, we added two comparative methods, including multi-sample self-consistency and iterative optimization. The results are shown in Table \ref{tab:compute_match}. It can be seen that PrismAgent still achieves the best performance across all three datasets. This validates that the performance gains of our method originate from the structured role decomposition and multi-agent collaboration design, rather than a mere increase in computational resources.

\subsection{Different Encoder}
\label{appendix:differebt encoder}

We conducted experiments using various encoders such as VisualBERT \cite{LiYatskarYinHsiehChang_2019} and BLIP-2 \cite{Chen_2023_blip}. The results are show in Table \ref{tab:encoders}. It can be seen that our method can acheive satisfactory performance improvements on all datasets. This illustrates that our framework is robust to different retrieval encoders architectures.

\subsection{Computational Overhead}
\label{appendix:computational overhead}
To clarify the computational cost, we compared the average inference time of PrismAgent with two baselines, MIND and single-step prompt, and the experimental results are presented in Table \ref{tab:time}
It can be seen that the multi-agent design and retrieval process inevitably introduce additional computational steps compared with single-step inference models, but in the comparison with another agent-based method MIND, our computational overhead is comparable. While our approach inevitably increases computational overhead, it achieves nearly a 10-point improvement in F1 score over single-step methods. This significant performance gain justifies the additional cost. In real-world applications, the computational time can be mitigated through multi-GPU parallelization or more capable hardware. We will further explore optimization strategies to reduce the model’s time overhead in future work.

\subsection{Reference Scale}
\label{appendix:reference scale}

To investigate the impact of different reference set sizes on PrismAgent, we conducted additional ablation experiments and the results are summarized in Table \ref{tab:reference_scale}. As the size of the reference pool increases, performance improves accordingly. Nevertheless, even with a small-scale reference pool, our method still achieves satisfactory results. This suggests that our approach may effectively adapt to emerging memes by simply expanding the reference pool, without requiring additional fine-tuning.

\subsection{Core Representation}
\label{appendix:core representation}

To investigate the impact of core representation on PrismAgent, we conducted additional ablation experiments and the results are summarized in Table \ref{tab:ablation_core}. It can be seen that the absence of the core representation results in obvious performance degradation. This is likely because the core representation serves as a condensed semantic features that captures the essential evidence and combines the information form the original and the rewritten memes. Without it, the model must rely on scattered or less structured information, which weakens its reasoning effectiveness and decision reliability.

\end{document}